\pdfoutput=1

\documentclass[11pt]{article}

\usepackage[]{acl}

\usepackage{times}
\usepackage{latexsym}
\usepackage{graphicx}
\usepackage{subfigure}
\usepackage{tcolorbox}
\usepackage[T1]{fontenc}

\usepackage[utf8]{inputenc}

\usepackage{microtype}
\usepackage{hyperref}

\title{UniGDD: A Unified Generative Framework for Goal-Oriented Document-Grounded Dialogue
\thanks{\hspace{1mm} The work described in this paper is substantially supported by a grant from the Research Grant Council of the Hong Kong Special Administrative Region, China (Project Code: 14200620).}
}

\author{Chang Gao, Wenxuan Zhang, and Wai Lam \\
  The Chinese University of Hong Kong \\
  \texttt{\{gaochang,wxzhang,wlam\}@se.cuhk.edu.hk} \\}

\begin{document}
\maketitle


\begin{abstract}
The goal-oriented document-grounded dialogue aims at responding to the user query based on the dialogue context and supporting document. Existing studies tackle this problem by decomposing it into two sub-tasks: knowledge identification and response generation. However, such pipeline methods would unavoidably suffer from the error propagation issue. This paper proposes to unify these two sub-tasks via sequentially generating the grounding knowledge and the response. We further develop a prompt-connected multi-task learning strategy to model the characteristics and connections of different tasks and introduce linear temperature scheduling to reduce the negative effect of irrelevant document information. Experimental results demonstrate the effectiveness of our framework.
\end{abstract}

\section{Introduction}

\label{intro}
Recent years have seen significant progress in goal-oriented dialogues \cite{bordes2017learning,  pmlr-v70-wen17a, wu-etal-2019-transferable, SimpleTOD, soloist}, which aim at assisting end users in accomplishing certain goals via natural language interactions. However, due to the lack of external knowledge, most goal-oriented dialogue systems are restricted to providing information that can only be handled by given databases or APIs \cite{kim-etal-2020-beyond} and completing certain tasks in a specific domain such as restaurant booking.
To address this challenge, goal-oriented document-grounded dialogue has been proposed to leverage external documents as the knowledge source to assist the dialogue system in satisfying users' diverse information needs \cite{feng-etal-2020-doc2dial, wu2021dialki}.

\begin{figure}[tb]
  \centering
  \includegraphics[width=\linewidth]{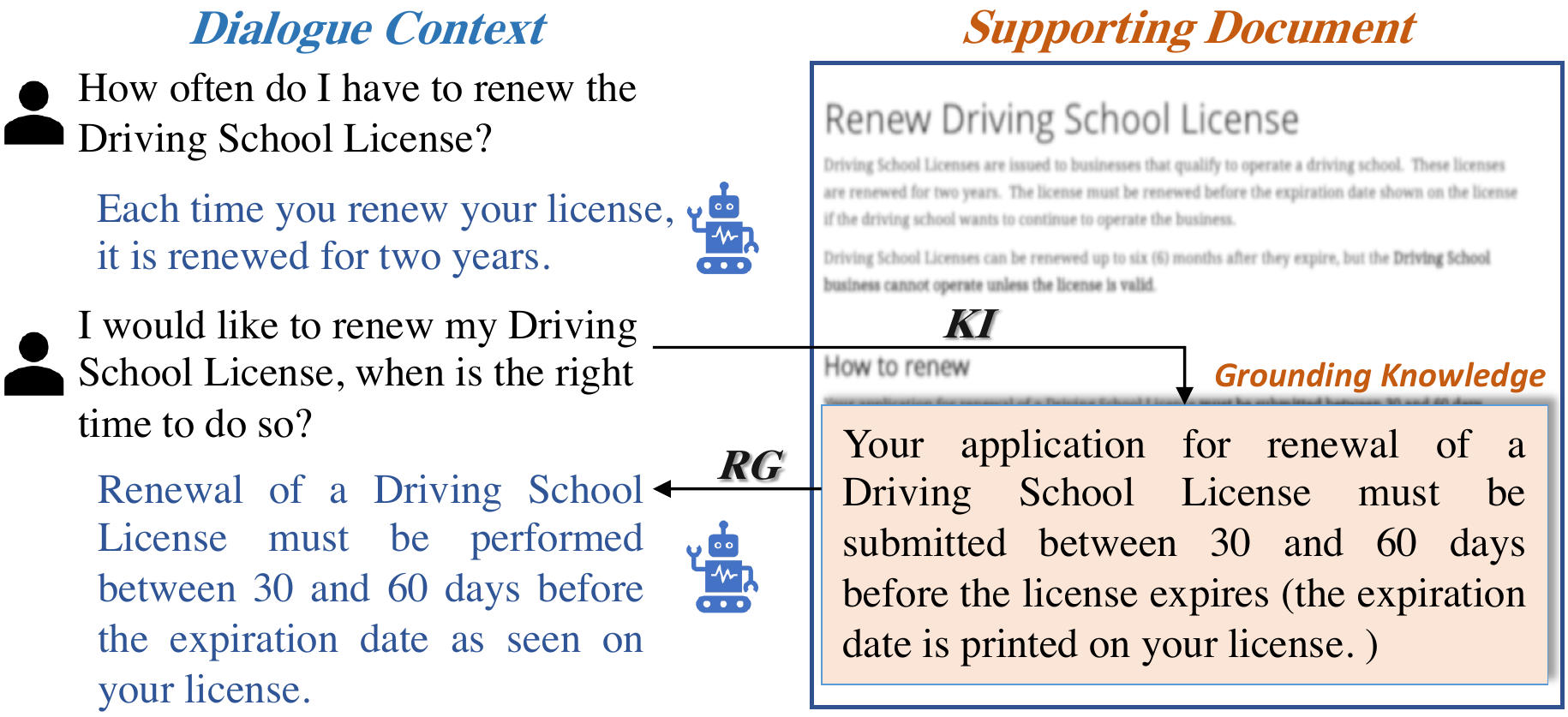} 
  \caption{An example of the goal-oriented document-grounded dialogue problem.}
  \label{fig:task}
\end{figure}

As shown in Figure \ref{fig:task}, the goal-oriented document-grounded dialogue problem is commonly formulated as a sequential process including two sub-tasks: knowledge identification (KI) and response generation (RG) \cite{feng-2021-dialdoc}. Given the dialogue context and supporting document, knowledge identification aims to identify a text span in the document as the grounding knowledge for the next agent response, which is often formulated as a conversational reading comprehension task \cite{feng-2021-dialdoc, wu2021dialki}. Response generation then aims at generating a proper agent response according to the dialogue context and the selected knowledge. Therefore, one straightforward solution for this problem is to use two models to conduct KI and RG in a pipeline manner \cite{daheim-etal-2021-cascaded,  kim-etal-2021-document, xu-etal-2021-caire, chen-etal-2021-building, scir}.
However, such pipeline methods fail to capture the interdependence between KI and RG. As a result, error propagation is a serious problem. The problem is more pronounced in low-resource scenarios, where accurate knowledge identification is difficult due to limited data, making it harder to generate appropriate responses.

To address the aforementioned issue, we propose a \textbf{Uni}fied generative framework for \textbf{G}oal-oriented \textbf{D}ocument-grounded \textbf{D}ialogue (\textbf{UniGDD}).  Given the dialogue context and associated document, instead of treating KI and RG as two separate processes, we tackle them simultaneously via sequentially generating the grounding knowledge and the agent response. Therefore, the inherent dependencies between these two sub-tasks can be naturally modeled. On one hand, the generation of the agent response depends not only on the dialogue context and external document but also on the identified knowledge, forcing the model to focus on the specific knowledge. On the other hand, the generation of the grounding knowledge receives the supervision signal from the agent response when training, leading to more accurate knowledge identification.

Although KI and RG can be unified with the proposed generative method, they have different characteristics. Generating the grounding knowledge is similar to copying appropriate sentences from the document, while generating the response needs more effort to make the response coherent with the dialogue and consistent with the grounding knowledge. Therefore, in addition to the main task that uses the concatenation of the grounding knowledge and response as the target sequence, we introduce the generation of the grounding knowledge and the generation of the response as two auxiliary tasks in the same framework to force the model to capture their characteristics so as to perform well on them as well. Moreover, inspired by the recent success in prompt learning for pre-trained models \cite{li-liang-2021-prefix, prompttuning, liu2021pretrain}, we design prompts for these three tasks to guide the model on what to generate for each task. These prompts can naturally connect these tasks via indicating the model that each auxiliary task aims to generate a part of the target sequence of the main task. Through this prompt-connected multi-task learning strategy, the model can capture the characteristics of different tasks as well as exploit the connections between them. 

In addition, for a particular user query in the goal-oriented dialogue, the selected knowledge and generated response need to be specific, while the generation conditions on a relatively long document. Thus, much information in the input document is irrelevant.  To tackle this problem, we introduce linear temperature scheduling to make the attention distribution to the input document gradually sharper during the training process in order to enable the model to learn to pay more attention to the relevant content. 

Our contributions are summarized as follows: (1)
 We propose a unified generative framework for the goal-oriented document-grounded dialogue. 
(2) We develop a prompt-connected multi-task learning strategy to exploit the characteristics and connections of different tasks and introduce linear temperature scheduling to enable the model to pay more attention to relevant information.
(3) Our framework advances state-of-the-art methods on the concerned task, especially in low-resource scenarios.

\section{Our UniGDD framework}

\begin{figure}[tb]
  \centering
  \includegraphics[width=\linewidth]{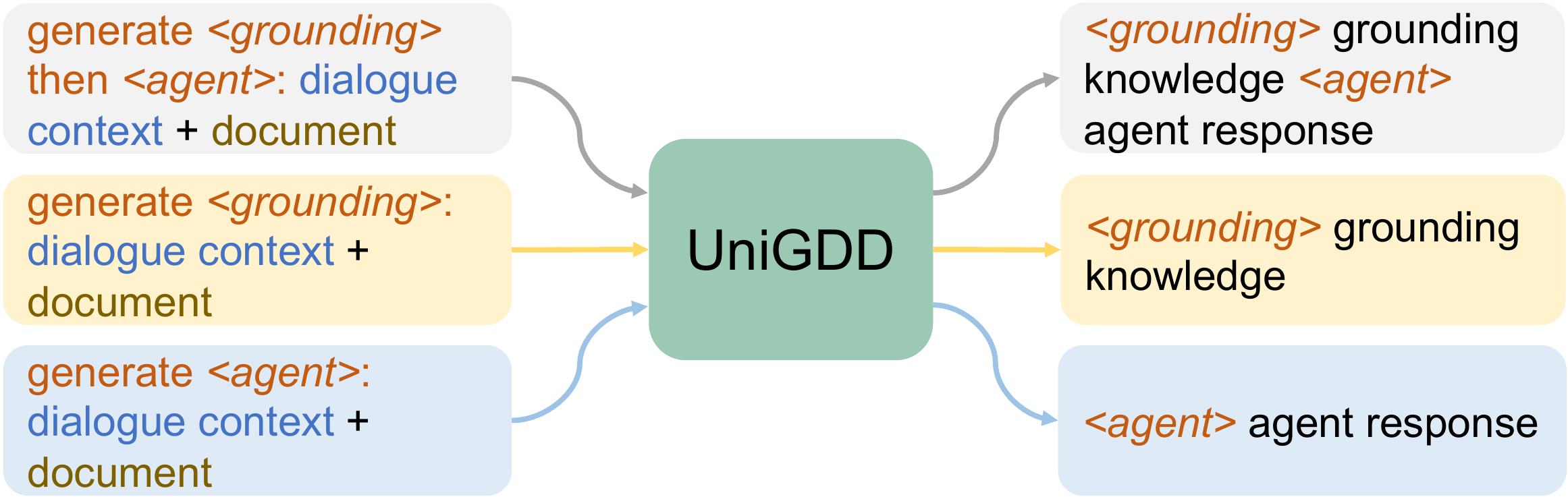}  
  \caption{Overview of our framework.}
  \label{fig:model}
\end{figure}

UniGDD is a multi-task generative framework for the goal-oriented document-grounded dialogue problem.

\textbf{Main Task} Given the dialogue context $C = (u_1, a_1, \dots, u_{t-1}, a_{t-1}, u_t)$ and grounding document $D$, where $u_i$ is the $i$-th user utterance and $a_i$ is the $i$-th agent utterance, our main task aims to generate the target sequence $Y=(k_t,a_t)$, where $k_t$ is the grounding knowledge from $D$ and $a_t$ is the response to $u_t$. 
Specifically, for the example in Figure \ref{fig:task},  the input and output of the main task are as follows:
\newtcolorbox{mybox}{colframe = black!75!black}
\begin{mybox}
 \textcolor{red}{Input:} \textcolor{orange}{generate \textsl{<grounding>} then \textsl{<agent>}:} \textcolor{blue}{ \textsl{<user>} I would like to renew ... ? \textsl{<agent>} Each time you ... \textsl{<user>} How often do ... ?}  \textcolor{brown}{\textsl{<title>} Renew Driving School License \textsl{</title>} ... Your application for renewal ... } \\
 \textcolor{red}{Output:} \textcolor{orange}{\textsl{<grounding>}} Your application for ... \textcolor{orange}{\textsl{<agent>}} Renewal of a Driving ...
\end{mybox}
\noindent We use different special tokens to identify different elements in the input and output. For example, we add \textsl{"<user>"} in front of each user utterance, \textsl{"<agent>"} in front of each agent utterance, and \textsl{"<grounding>"} in front of the grounding knowledge. The prompt "generate \textsl{<grounding>} then \textsl{<agent>}:" is added to the dialogue context and supporting document to form the input and guide the model  to generate the grounding knowledge and the response in order. The input-to-target generation can be modeled with a pre-trained encoder-decoder model $\mathcal{M}: (C,D,TP)\rightarrow (k_t, a_t)$ such as T5 \cite{T5}, where $TP$ is the task prompt.

\textbf{Prompt-Connected Multi-Task Learning}
We introduce two auxiliary tasks to steer our framework to model the respective characteristics of knowledge identification and response generation. Given the dialogue context $C$ and grounding document $D$, these two tasks aim to generate  the grounding knowledge $k_t$ and the response $a_t$ with the same model $\mathcal{M}$. As depicted in Figure \ref{fig:model}, we construct prompts "generate \textsl{<grounding>}:" and  "generate \textsl{<agent>}:" for them. These prompts indicate the model that the goals of the two auxiliary tasks are to generate the first part and the second part of the target sequence of the main task, respectively. As a result, the connections between different tasks are naturally modeled. Instead of using discrete language phrases, we randomly initialize the embeddings of those special tokens in the prompts and train them end-to-end to better encode the characteristics and connections of these tasks.

\textbf{Linear Temperature Scheduling}
For a specific user query in the dialogue, many document contents are actually irrelevant. To force the model to pay less attention to the irrelevant parts, we propose a linear temperature scheduling strategy to make the attention distribution of cross-attention gradually sharper during the training process. 
Specifically, we design the \texttt{softmax} function in the cross-attention module of each decoder layer as follows:
\begin{equation}
    a_i = \frac{\exp{(z_i/\tau)}}{\sum_{j}\exp{(z_j/\tau)}}
\end{equation}
\begin{equation}
    \tau = (\tau_e - \tau_s) \frac{S_c}{S_{total}} + \tau_s
\end{equation}
where $a_i$ is the attention weight for the $i$-th input token, $z_i$ is the logit for the $i$-th input token, $S_c$ is the current training step, $S_{total}$ is the total training steps, $\tau_s$ and $\tau_e$ are the starting and ending temperature respectively, $\tau_e < \tau_s$, and $0 < \tau_e < 1$. 
Compared with the original cross-attention module, the ending temperature $0 < \tau_e < 1$ leads to a sharper attention distribution, giving more attention weight to the relevant content.

\textbf{Training} \ The model is trained with a maximum likelihood objective. Given the training example $e=(C,D,TP,Y)$, the objective $L_\theta$ is defined as
\begin{equation}
\mathcal{L}_{\theta}=-\sum_{i=1}^{n} \log P_{\theta}\left(Y_{i} \mid Y_{<i}, C, D, TP\right)
\end{equation}
where $\theta$ is the model parameters, $TP$ is the task prompt, $Y$ is the target sequence, and $n$ is the length of $Y$. We mix the data of the main task and two auxiliary tasks for training.

\textbf{Inference} After training, for each pair of dialogue context and document $(C, D)$, we generate the target sequence of the main task for obtaining the grounding knowledge $k_t$ and the response $a_t$.

\section{Experiments}

\begin{table}
\centering
\begin{tabular}{lcc}
\hline
\textbf{Models} & \textbf{EM} & \textbf{F1} \\
\hline
BERTQA & 42.2 & 58.1 \\
BERT-PR-large  & 56.3 & 70.8 \\
RoBERTa-PR-large  & 65.6 & 77.3\\
Multi-Sentence  & 59.5 & 68.8 \\
DIALKI ($\mathcal{L}_{next}$ only) & 60.4 & 71.2 \\
DIALKI   & 65.9 & 74.8 \\
\hline
UniGDD-base & 65.6 & 76.8 \\
UniGDD-large & \textbf{66.9} & \textbf{77.5} \\
\hline
\end{tabular}
\caption{Results on knowledge identification.}
\label{tab:KI}
\end{table}

\begin{table}
\centering
\begin{tabular}{lc}
\hline
\textbf{Models} & \textbf{BLEU} \\
\hline
DIALKI+BART-base   & 25.8 \\
RoBERTa-PR-large+BART-base & 39.6 \\
RoBERTa-large+T5-base & 40.7 \\
\hline
UniGDD-base & 42.8 \\
UniGDD-large & \textbf{42.9} \\
\hline
\end{tabular}
\caption{Results on response generation.}
\label{tab:RG}
\vspace{-0.0cm}
\end{table}

\subsection{Experimental Setup}
\quad\textbf{Dataset} We conduct experiments on the goal-oriented document-grounded dialogue dataset Doc2Dial \cite{feng-2021-dialdoc}, which is adopted by the DialDoc21 shared task\footnote{ \href{https://github.com/doc2dial/sharedtask-dialdoc2021}{https://github.com/doc2dial/sharedtask-dialdoc2021}}. It contains 3,474 dialogues with 44,149 turns for training and 661 dialogues with 8539 turns for evaluation\footnote{Since we cannot access the test set, we report results on the development set for comparison.}. 

\textbf{Evaluation Metrics} Following \citet{feng-2021-dialdoc}, we use Exact Match (EM) and token-level F1  for knowledge identification and BLEU \cite{bleu, sacrebleu} for response generation.

\textbf{Baselines} For knowledge identification, we compare UniGDD with several strong baselines, including BERTQA \cite{devlin-etal-2019-bert}, BERT-PR \cite{daheim-etal-2021-cascaded}, RoBERTa-PR \cite{daheim-etal-2021-cascaded}, Multi-Sentence \cite{wu2021dialki}, and DIALKI \cite{wu2021dialki}. These models formulate knowledge identification as the machine reading comprehension task and extract the grounding span from the document.  For response generation, we compare UniGDD with several pipeline methods, including DIALKI+BART \cite{wu2021dialki} that uses DIALKI to conduct knowledge identification, followed by BART \cite{lewis-etal-2020-bart} to conduct response generation and RoBERTa-PR+BART \cite{daheim-etal-2021-cascaded}.  We also build a strong baseline model RoBERTa+T5 which uses the same pre-trained generative model as ours.

\textbf{Implementation Details} We report results of UniGDD with two model sizes: UniGDD-base and UniGDD-large, which are initialized with pre-trained T5-base and T5-large models \cite{T5}, respectively. We adopt the implementation from Hugging Face Transformers \cite{wolf-etal-2020-transformers}.  We set the max input length to 2560. Any sequence over 2560 tokens will be truncated. For training, we use the AdamW \cite{adamw} optimizer with an initial learning rate of $10^{-4}$ and a linear learning rate decay scheduler. We train 10 epochs for single-task learning and 5 epochs for multi-task learning. For decoding, we use beam search, and the beam size is 2. For linear temperature scheduling, we set the starting temperature $\tau_s=1$ and choose the best ending temperature from \{0.5, 0.6, 0.7, 0.8, 0.9\}. 
For our constructed baseline RoBERTa+T5 for response generation, we use RoBERTa-large and T5-base and adopt the implementation from the DialDoc21 shared task.

\subsection{Results}
The results on knowledge identification and response generation are shown in Table \ref{tab:KI} and Table \ref{tab:RG}, respectively. Our UniGDD framework outperforms all the baselines on two sub-tasks. On the knowledge identification task, UniGDD-base can obtain comparable results to previous state-of-the-art methods. With a larger model size, UniGDD-large achieves new state-of-the-art performance.
On the response generation task, UniGDD obtains a marked improvement over all pipeline methods. This verifies our assumption that our unified generative framework can alleviate the error propagation problem of pipeline approaches.

\textbf{Effect of Prompt-Connected Multi-task Learning (PCMTL) and Linear Temperature Scheduling (LTS)} To verify the effectiveness of PCMTL and LTS, we first remove PCMTL (i.e., training with the main task only), and the performance of UniGDD-base on two tasks decreases to 65.2 EM, 76.3 F1, and 42.3 BLEU, showing that PCMTL endows the model with the ability of modeling the characteristics and connections of different tasks and achieving better generation.
Further removing LTS, the performance drops to 64.7 EM, 76.0 F1, and 41.7 BLEU. This indicates that LTS can guide the model to pay more attention to relevant content during generation and bring improvements on two sub-tasks.

\textbf{Effect of Connected Prompts (CP)} To examine whether CP can capture the connections of different tasks, we use an alternative approach that employs task-independent prompts "<Task1>:", "<Task2>:", and "<Task3>:" to specify each task for comparison. As in the case of CP, we randomly initialize the embeddings of these three special tokens. With these prompts, UniGDD-base obtains 64.9 EM, 76.2 F1, and 42.3 BLEU, which performs worse than using CP. This indicates that CP enables the model to take advantage of the connections between the three tasks. 

\begin{figure}[tb]
\centering
\subfigure[EM]{
\centering
\begin{minipage}[t]{0.5\linewidth}
\centering
\includegraphics[width=1.5in]{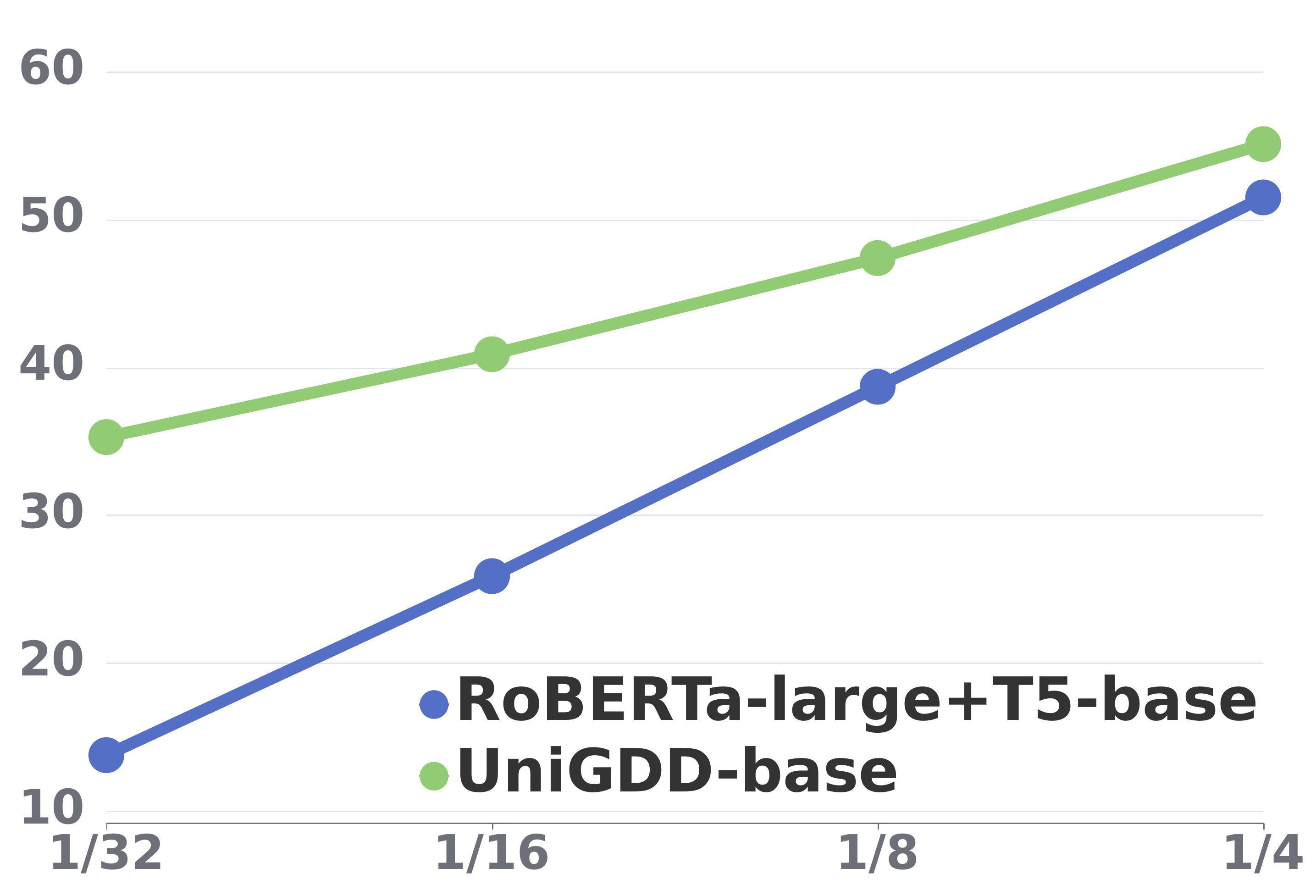}
\end{minipage}%
}%
\subfigure[BLEU]{
\centering
\begin{minipage}[t]{0.5\linewidth}
\centering
\includegraphics[width=1.5in]{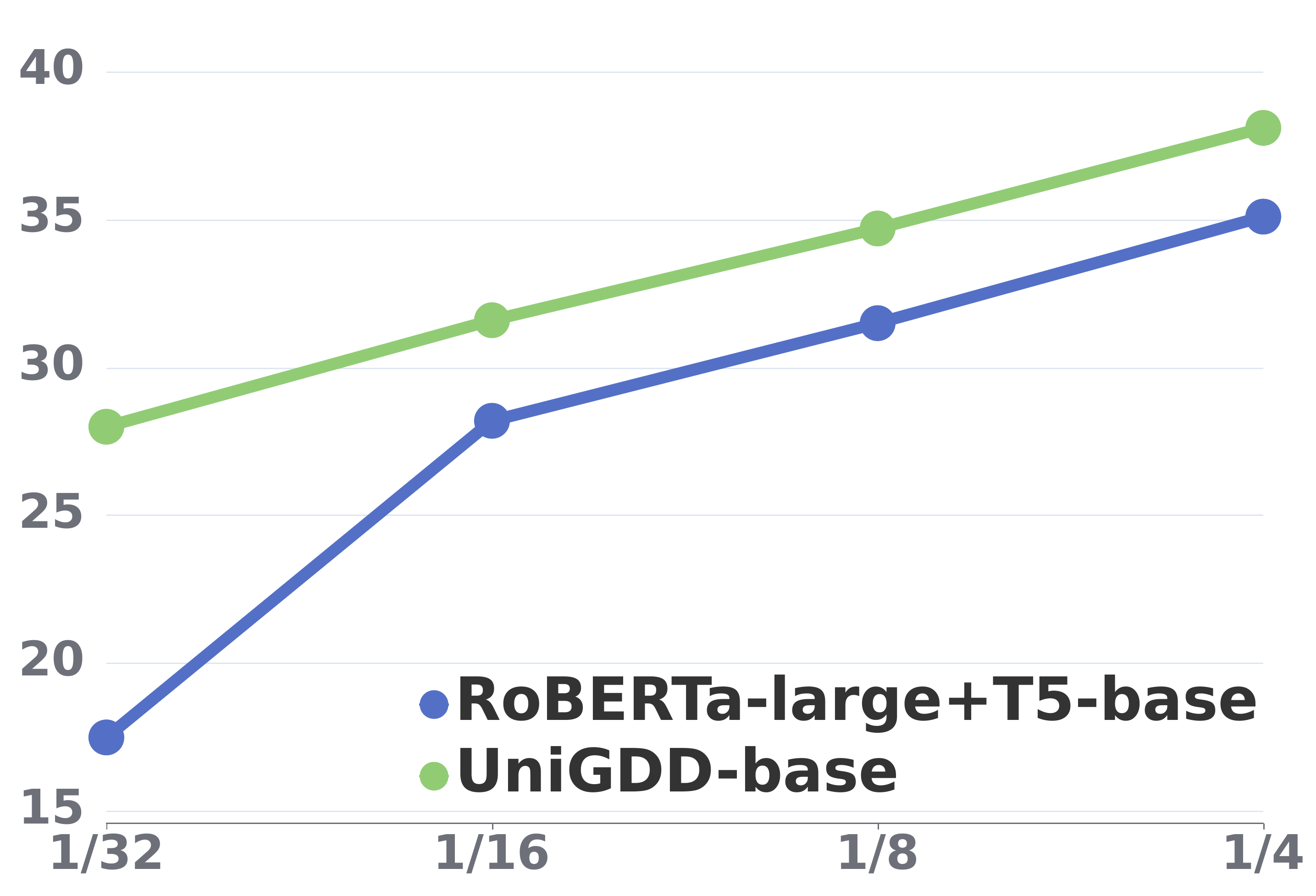}
\end{minipage}%
}
\caption{Experimental results on knowledge identification and response generation in low-resource scenarios}
\label{fig:low}
\end{figure}

\textbf{Low-Resource Setting}
To evaluate the model in low-resource scenarios, we randomly shuffle the training set and then take 1/32, 1/16, 1/8, and 1/4 of the data for training.
Figure \ref{fig:low} shows the results of UniGDD-base and the best-performing pipeline baseline RoBERTa-large+T5-base on the four low-resource training splits. Generally, our framework performs substantially better than the pipeline method on both tasks. Particularly, when there is only 1/32 training data, UniGDD-base obtains more than 20 and 10 absolute points improvement over the pipeline approach on EM and BLEU, respectively. 

\begin{figure}[htb]
  \centering
  \includegraphics[width=\linewidth]{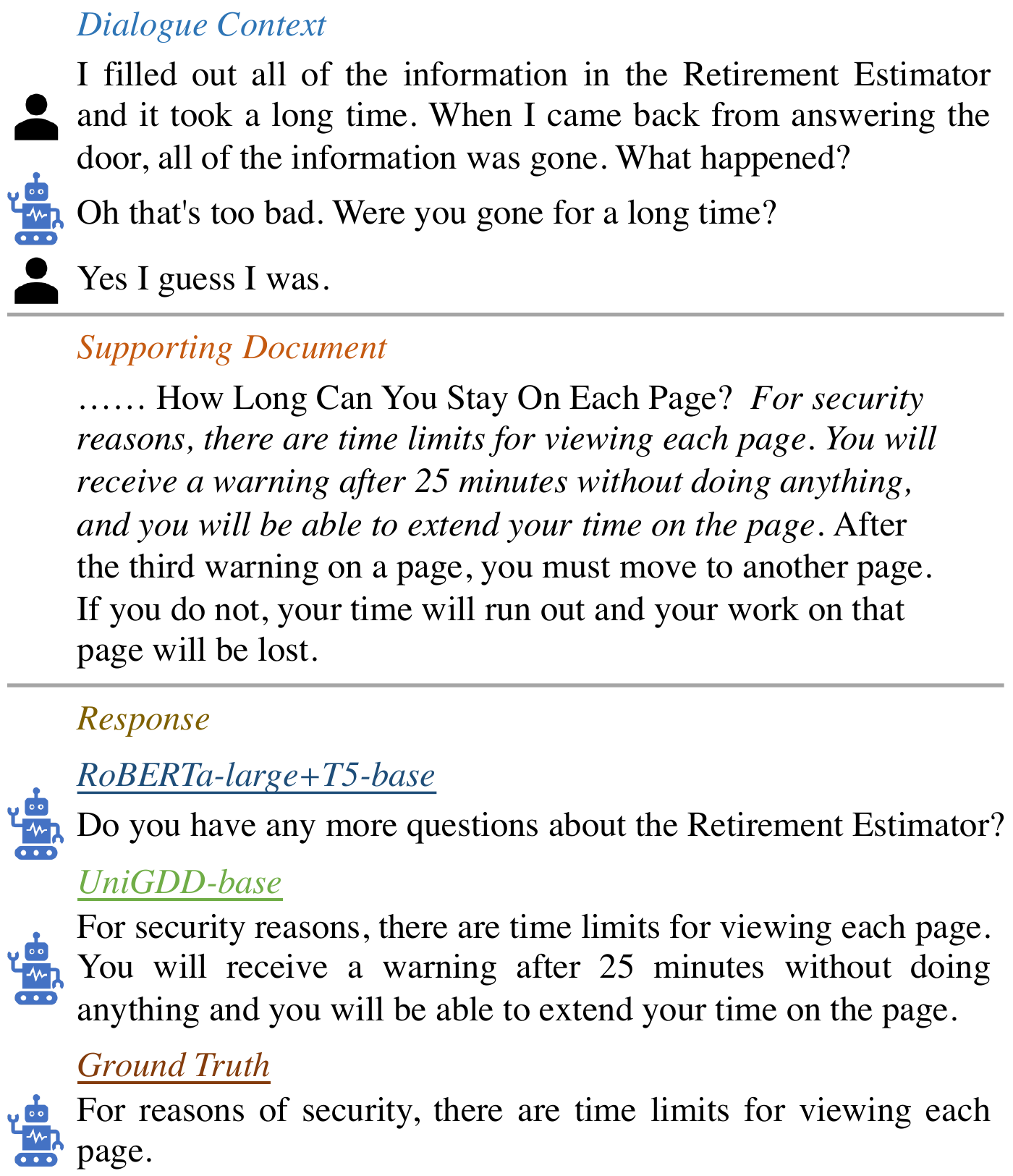}
  \caption{A case from the development set.}
  \label{fig:case}
\end{figure}

\textbf{Case Study}
Figure \ref{fig:case} shows a real case including the dialogue context, supporting document, and the responses generated by the pipeline method and our proposed UniGDD framework. It can be observed that our framework identifies accurate knowledge from the supporting document and thus provides a proper and informative response about the reasons for the problem the user encounters. In contrast, the pipeline method only gives a relatively general response that is not suitable in this case.

\subsection{Human Evaluation}

We randomly sample 100 evaluation instances. For each instance, given the dialogue context and grounding document, three human annotators are asked to conduct a pairwise comparison between the response generated by UniGDD-base and the one generated by the pipeline baseline RoBERTa-large+T5-base in terms of two aspects: (1) \textit{Relevance}: which response is more relevant and appropriate to the user query? (2) \textit{Informativeness}: which response is more informative? Results are shown in Table \ref{tab:Human}. 
Compared with the pipeline method, our framework can reduce error propagation, resulting in more relevant and appropriate responses.  Moreover, our framework has a clear advantage over the baseline in terms of  Informativeness since it can utilize rich document context during the generation.

\begin{table}[htb]
\centering
\begin{tabular}{lccc}
\hline
& \textbf{Win} & \textbf{Tie} & \textbf{Lose} \\
\hline
Relevance   & 26 & 64 & 10 \\
Informativeness & 23 & 69 & 8 \\
\hline
\end{tabular}
\caption{UniGDD-base vs RoBERTa-large+T5-base. The numbers indicate how many instances there are in each case.}
\label{tab:Human}
\vspace{-0.2cm}
\end{table}

\section{Conclusion}
Our UniGDD framework unifies knowledge identification and response generation and models their characteristics via a multi-task generative modeling strategy. Both automatic evaluation and human evaluation demonstrate the effectiveness of our framework.


\end{document}